\documentclass[conference]{IEEEtran}
\IEEEoverridecommandlockouts
\usepackage[T1]{fontenc}
\usepackage{cite}
\usepackage{amsmath,amssymb,amsfonts}
\usepackage{bbding}
\usepackage{graphicx}
\usepackage{amssymb}
\usepackage{hyperref}
\usepackage{amsmath}
\usepackage{tabularx}
\usepackage{booktabs}  
\usepackage[table, dvipsnames]{xcolor}
\usepackage{subcaption}
\usepackage{caption}
\usepackage{svg}
\usepackage{tcolorbox}
\definecolor{lightgray}{gray}{0.9}
\definecolor{darkgreen}{rgb}{0,0.5,0}

\newcommand{\greencheck}{\textcolor{darkgreen}{\CheckmarkBold}}
\newcommand{\redcross}{\textcolor{red}{\XSolidBrush}}

\let\emptyset\varnothing

\newcolumntype{C}[1]{>{\centering\arraybackslash}p{#1}}
\newcolumntype{M}{>{\hsize=0.5\hsize\raggedright\arraybackslash}X}

\def\BibTeX{{\rm B\kern-.05em{\sc i\kern-.025em b}\kern-.08em
    T\kern-.1667em\lower.7ex\hbox{E}\kern-.125emX}}
\begin{document}

\title{Catastrophic Forgetting Resilient One-Shot Incremental Federated Learning}



\author{
\IEEEauthorblockN{ Obaidullah Zaland, Zulfiqar Ahmad Khan, and Monowar Bhuyan}
\IEEEauthorblockA{\textit{Department of Computing Science},
\textit{Umeå University},
Umeå, SE-90187, Sweden,\\
 E-mail: \{ozaland, zakhan, monowar\}@cs.umu.se}
\thanks{This work was partially supported by the Wallenberg AI, Autonomous Systems and Software Program (WASP) funded by the Knut and Alice Wallenberg Foundation via the WASP NEST project “Intelligent Cloud Robotics for Real-Time Manipulation at Scale.” The computations and data handling essential to our research were enabled by the supercomputing resource Berzelius provided by the National Supercomputer Centre at Linköping University and the gracious support of the Knut and Alice Wallenberg Foundation.}
}

\maketitle

\begin{abstract}

Modern big-data systems generate massive, heterogeneous, and geographically dispersed streams that are large-scale and privacy-sensitive, making centralization challenging. While federated learning (FL) provides a privacy-enhancing training mechanism, it assumes a static data flow and learns a collaborative model over multiple rounds, making learning with \textit{incremental} data challenging in limited-communication scenarios. This paper presents One-Shot Incremental Federated Learning (OSI-FL), the first FL framework that addresses the dual challenges of communication overhead and catastrophic forgetting. OSI-FL communicates category-specific embeddings, devised by a frozen vision-language model (VLM) from each client in a single communication round, which a pre-trained diffusion model at the server uses to synthesize new data similar to the client's data distribution. The synthesized samples are used on the server for training. However, two challenges still persist: i) tasks arriving incrementally need to retrain the global model, and ii) as future tasks arrive, retraining the model introduces catastrophic forgetting. To this end, we augment training with Selective Sample Retention (SSR), which identifies and retains the top-p most informative samples per category and task pair based on sample loss. SSR bounds forgetting by ensuring that representative retained samples are incorporated into training in further iterations. The experimental results indicate that OSI-FL outperforms baselines, including traditional and one-shot FL approaches, in both class-incremental and domain-incremental scenarios across three benchmark datasets.
\end{abstract}

\section{Introduction}

\begin{figure}[t]
  \centering
  \begin{subfigure}{0.48\columnwidth}
    \centering
    \includegraphics[width=\linewidth]{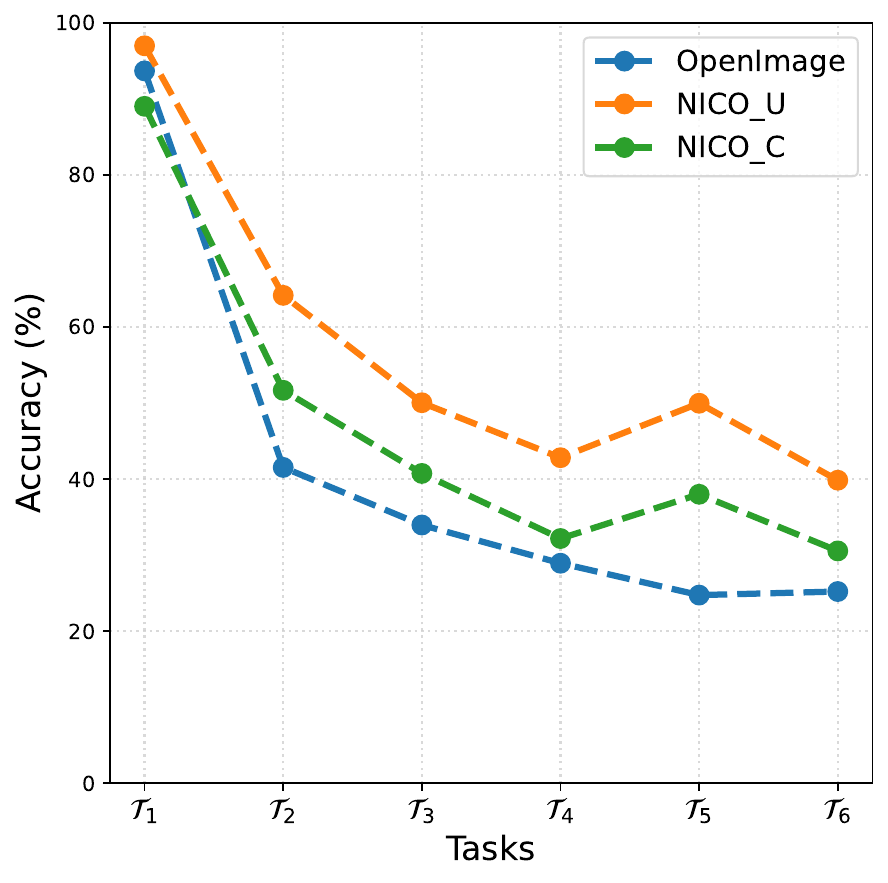}
    \caption{FedAvg}
    \label{fig:fedavg-sub}
  \end{subfigure}\hfill
  \begin{subfigure}{0.48\columnwidth}
    \centering
    \includegraphics[width=\linewidth]{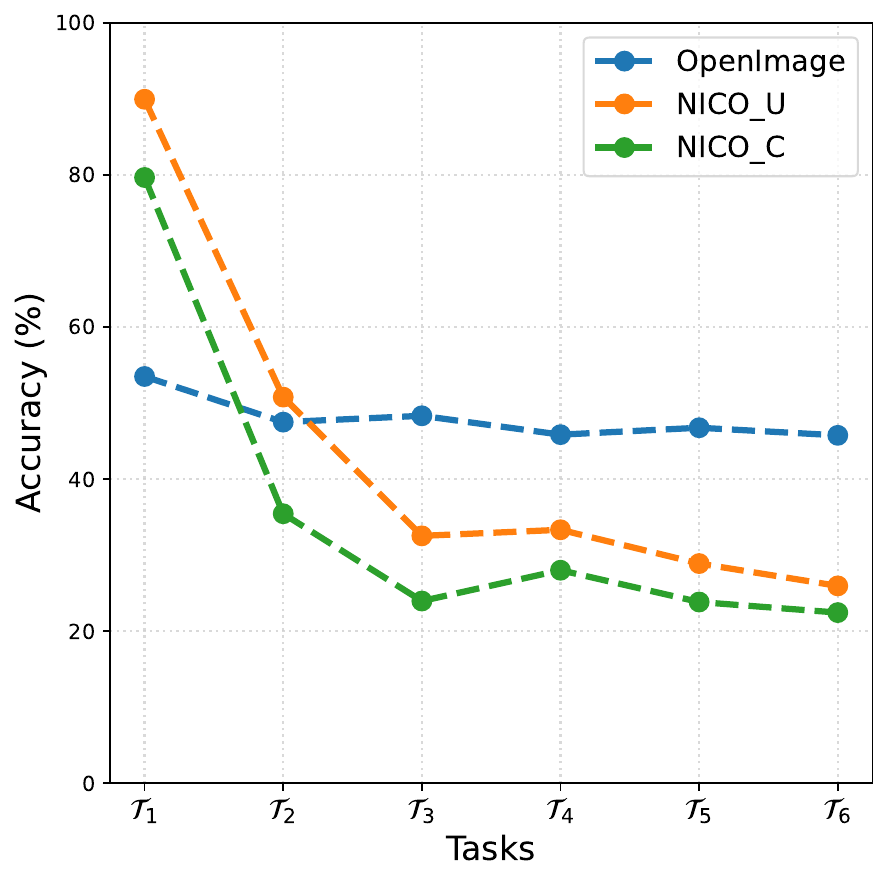}
    \caption{OSCAR-IL}
    \label{fig:oscaril-sub}
  \end{subfigure}
  \caption{Motivation behind OSI-FL. One-shot FL approaches degrade in performance similar to traditional FL approaches as tasks arrive incrementally, specially when the tasks differ significantly.}
  \label{fig:both-algos}
\end{figure}

\begin{figure*}
    \centering
    \includegraphics[width=\linewidth]{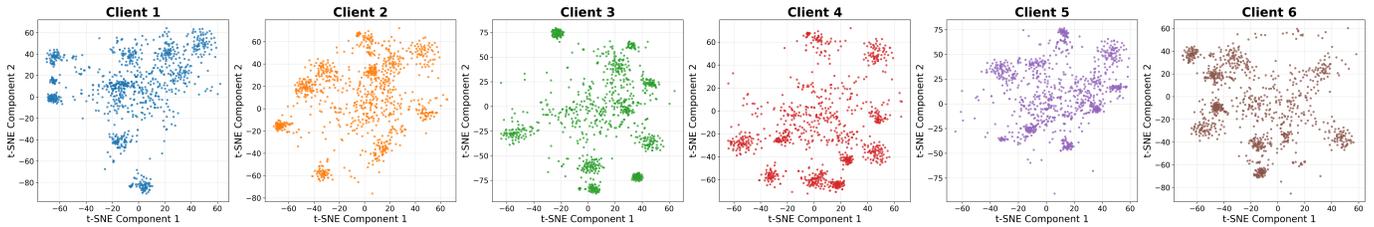}
    \caption{T-SNE data visualization for multiple tasks, where each task contains a subset of the total classes, in an incremental FL setup (NICO Common dataset). As depicted, the distribution of data varies significantly among tasks, and as these tasks arrive incremetnally, learning the most recent task affects the knowledge about previous tasks (catastrophic forgetting).}
    \label{fig:tsnescatter}
\end{figure*}

Conventional machine learning (ML) training involves data centralization, which, at big-data scale, often becomes infeasible as datasets grow and span institutions, raising concerns about data ownership, privacy, and regulatory compliance~\cite{zalandiconip}. Federated learning (FL)~\cite{mcmahan2017communication} extends the traditional machine learning (ML) paradigm to enable data storage at the source, while training a global model that captures knowledge across clients, usually through exchanging model updates. FL has thus found potential applications in various domains from healthcare~\cite{li2025challenges} to autonomous vehicles~\cite{fu2024secure}, due to its privacy-enhancing and ownership-preserving qualities. Despite its benefits, FL assumes that the data distribution remains static across clients, and new data will not be introduced, which violates real-world settings where data flow is continuous~\cite{ifldata}. For instance, in medical diagnosis, datasets are continuously updated~\cite{iqbal2025hierarchical}. The continuous flow of data, especially when the distributions of arriving tasks differ significantly (e.g. \textit{Figure} \ref{fig:tsnescatter}), affects the global model performance in traditional FL algorithms (see \textit{Figure} \ref{fig:both-algos}).


Incremental federated learning (IFL)~\cite{casado2023ensemble} has been introduced to bridge the gap in traditional FL by continuously learning the global model and adapting it to dynamic data flows at clients. IFL schemes introduce regularization~\cite{iflmodel2}, fine-tuning~\cite{yu2025efficient}, or replay buffers~\cite{ifldata2} to mitigate the data distribution shift. Incremental FL frameworks, however, face two significant challenges~\cite{hamedi2025federated}: \textbf { i) high communication costs}, since each distribution shift involves transmitting model gradients and updates across clients, and \textbf{ii) catastrophic forgetting}, where the global model's performance deteriorates on earlier tasks, and the global model adapts to the new task data. Current approaches, such as selective retraining~\cite{iflmodel}, elastic weight consolidation (EWC)~\cite{iflmodel2}, and data replay~\cite{ifldata,ifldata2}, while reducing the impact of catastrophic forgetting, overlook the communication costs incurred by constant client-server communication.

Recently, generative model-assisted one-shot federated learning (OSFL) has attracted increasing research interest~\cite{feddeo}, which leverages the generative power of pre-trained diffusion models (DMs) to reduce communication costs in FL, specifically limiting global communication to a single round between clients and the server. Current approaches utilize prompt engineering~\cite{feddeo}, data encoding~\cite{oscar}, or federated diffusion model training~\cite{mendieta2025navigating} to synthesize \textit{new data} that is similar to clients' data distributions at the server using pretrained DMs. The global model is then trained on \textit{synthesized data} generated by the DM. However, in line with existing research in traditional FL, OSFL approaches neglect the continual, dynamic nature of clients' data and assume it is static at the time of training a global model.

To address the limitations of both IFL and OSFL approaches, this work introduces One-Shot Incremental Federated Learning (\textbf{OSI-FL}), the first framework to integrate the incremental update process into a one-shot learning framework, thereby reducing communication cost to a single round per client. OSI-FL incorporates client-level category-specific data encodings, generated by a vision-language model (VLM) at the client to incrementally synthesize new data at the server using a classifier-free diffusion model (DM), inspired by OSCAR~\cite{oscar}, effectively mimicking the data distribution at the client. The data generation pipeline for the single-shot communication is similar to OSCAR. However, the description generator and the text encoder have been replaced with their lightweight counterparts to reduce the burden on individual clients. Due to the smaller size of data encodings and a single communication round for each client, OSI-FL significantly reduces the communication load for each client compared to traditional FL and incremental FL approaches.

Furthermore, to reduce the impact of catastrophic forgetting, OSI-FL incorporates selective sample retention (\textbf{SSR}) at the server. SSR eliminates the need to retrain the model on the full dataset from previous tasks when a new task arises. As each task arrives, OSI-FL trains the model on the data from the newly arrived task, alongside $p$ \textit{important} samples from each of the previous tasks, and identifies $p$ samples for the current task to be used for training on future tasks. In exemplar selection, SSR combines two strategies of class-balanced sampling and dominant gradient sampling to ensure both diversity and effectiveness of the retained samples. The $p$ exemplars for each task are stored on the server for replay, ensuring the model maintains its performance on previous tasks without retraining on the complete dataset. Furthermore, training with only $p$ samples rather than the whole dataset reduces the computational and memory cost at the server.

Overall, the contributions of this paper are listed as follows:
\begin{itemize}
    \item This work introduces the first one-shot incremental FL framework, which learns a global model with a single communication round per client. While there have been extensive studies of incremental FL, all of them rely on constant communication between clients and the server. 
    \item To mitigate catastrophic forgetting, this work incorporates a sample retention mechanism, where $k$ exemplars are selected from each task for data replay instead of retraining on the complete dataset as new tasks arrive. The sample selection is based on a combination of class-based and importance-based mechanisms, thereby improving the diversity of the selected samples.
    \item We provide a comprehensive analysis of our approach across three benchmarking datasets: NICO Common, NICO Unique, and OpenImage, against baseline FL, incremental FL, FedEWC, and one-shot FL, demonstrating its effectiveness in reducing communication and computation while preserving competitive performance. 
    \end{itemize}

\section{Related Work}

\subsubsection{Federated Learning (FL)} is a collaborative machine learning paradigm, where multiple clients train a joint global model over multiple communication rounds without sharing their private data~\cite{chen2024federated}. Typically, in each round, a \textit{central server} acts as a coordinator, distributes the global model to clients, who train it on their private data, before sending it back to the server. The server then aggregates the local client models to form an updated global model. While FedAvg~\cite{mcmahan2017communication} is the simplest aggregation method at the server, it faces challenges under heterogeneous data distribution (i.e., non-IID data). Subsequent approaches to tackle data heterogeneity, such as adding a proximal term in client optimization~\cite{li2020federated}, or adding control variates~\cite{karimireddy2020scaffold} to reduce client model drift. However, the aforementioned approaches require multiple communication rounds between clients and servers to achieve an optimal global model.

\subsubsection{One Shot Federated Learning (OSFL)~\cite{osfl}} learns a global model in a single communication round between the clients and the server~\cite{amato2025towards}. Existing FL approaches use distillation~\cite{dosfl} or auxiliary data generation~\cite{feddeo,oscar} to achieve competitive performance. Distillation-based OSFL approaches use the trained client model to distill the global model using a public dataset. Auxiliary data generation approaches train~\cite{dense}, or use pretrained generative models~\cite{oscar} (e.g., generative adversarial networks or diffusion models) to generate \textit{new data} that mimics the client data distribution for global model training. OSFL approaches considerably reduce communication costs in FL, but, like their conventional FL counterparts, they assume the data is static. The performance of these approaches deteriorates when tasks arrive incrementally, as they tend to forget knowledge from previous tasks.

\subsubsection{Incremental Federated Learning (IFL)} allows learning a global collaborative model under incrementally arriving tasks at clients~\cite{ifl}. IFL approaches balance learning and knowledge of the newly arrived task while retaining knowledge from previously learned tasks. The tasks may differ based on tasks, domains~\cite{ifldomain}, classes~\cite{iflclass}, or a combination of them. Existing approaches leverage data replay, regularization, or parameter space separation for IFL. Data replay-based approaches~\cite{ifldata,ifldata2} retain samples from previous tasks to mitigate catastrophic forgetting. Regularization-based approaches~\cite{iflmodel,iflmodel2} apply constraints on the loss function to prevent overwriting of the existing knowledge, and parameter space separation methods divide the parameter space amongst the sequence of tasks. Parameter space separation approaches~\cite{iflparameter} isolate the parameter space across tasks, leveraging distinct portions to retain knowledge from each task.

\begin{table}[!ht]
    \centering
    \caption{Symbols used}
    \begin{tabularx}{0.75\linewidth}{m{1cm}|X}
    \textbf{Symbol} & \textbf{Explanation}\\
    \hline
    $\mathcal{C}$ & Set of participating clients\\
    $\mathcal{T}$ & Set of all tasks\\
    $n$ & Number of participating clients\\
    $m$ & Number of tasks \\
    $\mathbf{D}_t$ & Data for task $t$\\
    $\mathbf{D}$ & Global Dataset\\
    $N_t$ & Number of samples for task $t$\\
    $\mathcal{Y}_t$ & Label space for task $t$ \\
    $\hat{\mathbf{D}}_t$ & Synthesized data for task $t$\\
    $\tilde{\mathbf{D}}_t$ & Retained synthesized samples of task $t$\\
    $p$ & Number of retained samples \\
    $e_{t,k}(i)^c$ & Embedding for class $k$, task $t$, client $c$ and image $i$\\
    $\mu_{t,k}^c$ & Average embedding of task $t$, class $k$, and client $c$\\
    $\theta_t$ & Model parameters after task $t$\\
    $\mathcal{E}_t$ & Set of retained samples for task $t$
    
    \end{tabularx}
    
    \label{tab:notations}
\end{table}

\section{Preliminaries}

\subsection{Incremental Federated Learning}\label{ifl}

Federated incremental learning considers $n$ clients, $\mathcal{C} = \{\mathcal{C}_1, \mathcal{C}_2, ..., \mathcal{C}_n\}$a series of $m$ tasks $\mathcal{T}=\{\mathcal{T}_1, \mathcal{T}_2,...,\mathcal{T}_m\}$, where for each task $T$, the data distribution $\mathbf{D}_t$ is fixed, but unknown in advance. \textit{Table} \ref{tab:notations} provides a list of notations used in this work and their definitions. Throughout this work, we assume that each client can only belong to one task from $\mathcal{T}$, while each task can have multiple clients associated with it. Each task contains $N_t$ number of samples for task $t$ denoted as $\mathbf{D}_t = \{x_t(i), y_t(i)\}$. The set of classes for each task $t$ is defined as $\mathcal{Y}_t$. In class incremental learning, for any two tasks $p,q$ where $p\neq q$, $\mathcal{Y}_p \cap \mathcal{Y}_q = \emptyset$, while in domain incremental learning, for any two tasks $p,q$ where $p\neq q$, $\mathcal{Y}_p$ and  $\mathcal{Y}_q$ may share some classes. The basic objective in FL is to learn an optimal model for all domains until the current domain, and can be written as:
\begin{equation}\label{eqn:ifl}
\theta_{t}
= \arg\min_{\theta}
\frac{1}{N_{t}}
\sum_{(x,y)\in \mathcal{D}_{t}}
\ell\bigl(f_{\theta}(x),y\bigr)
\end{equation}
Traditional FL fails to perform well under incremental learning settings, as the new task introduced usually leads to forgetting previously learned knowledge (catastrophic forgetting). Hence, more robust aggregation mechanisms are needed.

\subsection{Diffusion Models}
Diffusion models learn data distributions and generate data by gradually adding noise to samples in the forward pass and then denoising in the backward pass. Given, a data point $x_0$, noise is added over $Z$ steps via a markov chain  
\begin{equation}
q(x_z \mid x_{z-1}) = \mathcal{N}\bigl(x_z; \sqrt{1-\beta_z},x_{z-1},,\beta_z I\bigr)
\end{equation}
or in closed form
\begin{equation}
q(x_z \mid x_0) = \mathcal{N}\bigl(x_z; \sqrt{\alpha_z},x_0,(1-\bar\alpha_z)I\bigr),
\end{equation}
where $\alpha_z=1-\beta_z$, and $\bar\alpha_z=\prod_{s=1}^z \alpha_s$. The reverse process is learned by training a neural network $\epsilon_\theta$ to predict the added noise based on the objective
\begin{equation}
L(\theta) = \mathbb{E}{x_0,z,\epsilon}\Bigl[\bigl|\bigl|\epsilon - \epsilon_\theta(x_z,z,c)\bigr|\bigr|^2\Bigr]
\end{equation}
Iterative noise removal recovers samples that are consistent with the data distribution.

Classifier-free guidance steers generation without a separate classifier by training the model both conditionally and unconditionally (by dropping the condition $c$) with probability $P_{drop}$. In inference, the predictions are combined as
\begin{equation}
\hat\epsilon_\theta(x_z,z,c,w)  = \epsilon_\theta(x_z,z,\varnothing) + w\Bigl(\epsilon_\theta(x_z,z,c)-\epsilon_\theta(x_z,z,\varnothing)\Bigr)
\end{equation}
where $w\geq 1$ controls adherence to the condition $c$. The reverse step is then defined as 
\begin{equation}
p_\theta(x_{z-1}\mid x_z,c) = \mathcal{N}\bigl(x_{z-1};\mu_\theta(x_z,z,c),\Sigma_\theta(x_z,z,c)\bigr)
\end{equation}

\begin{table}[h]
\centering
\caption{Comparison between OSCAR and OSI-FL approaches}
\label{tab:oscar_vs_OSI-FL}
\begin{tabularx}{\linewidth}{m{1.3cm}|X|X}
\hline
\textbf{Criteria} & \textbf{OSCAR} & \textbf{OSI-FL} \\ 
\hline

\hline
Description Generation & OSCAR primarily utilizes the BLIP OPT pretrained model for caption generation, which is large, weighing around \textbf{5 GB}.  & In contrast, OSI-FL employs the GPT ViT pretrained model for captioning, which is compact, totaling around \textbf{0.9 GB}.  \\ 
\hline
Data Division & OSCAR mainly use domain-based data division, incorporating feature-skewed non-IID division to simulate a distributed setup. & OSI-FL adds a label-skewed, non-IID setup in which data is divided by label across clients. Although we also provide experimental data for a feature-skewed setup. \\ 
\hline
Learning Setup & OSCAR assume that all encodings from all clients are available at the start of the experiments, which is impractical in real-world settings. & OSI-FL incorporates a more realistic setup in which clients join the FL setup incrementally.\\ \hline

\hline
\end{tabularx}
\end{table}

\begin{figure*}
    \centering
    \includegraphics[width=\linewidth]{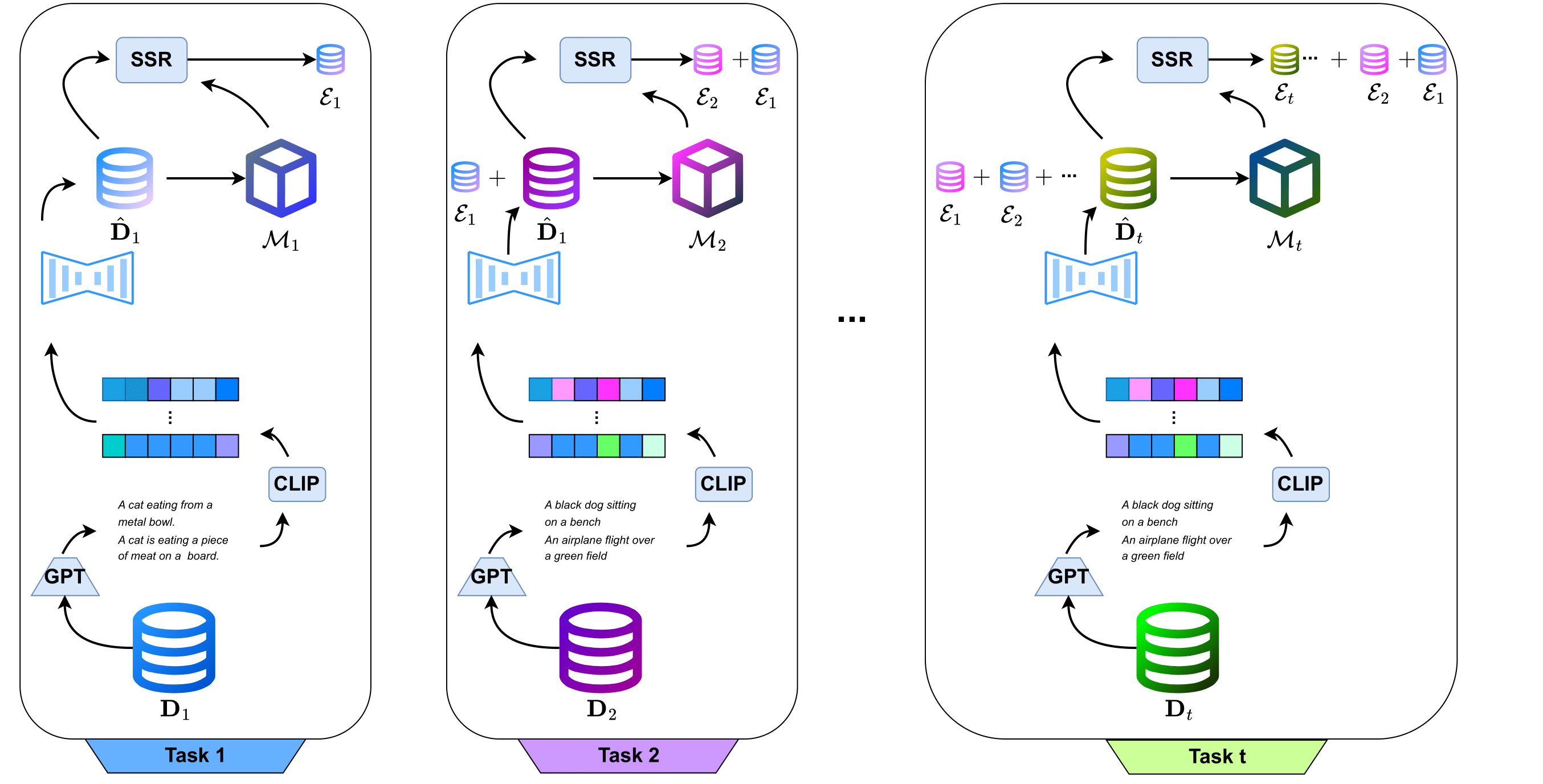}
    \caption{Overview of OSI-FL framework, as new tasks arrive.}
    \label{fig:osifl}
\end{figure*}

\section{System Design: One-Shot Incremental Federated Learning}

Similar to section \ref{ifl}, we consider $m$ tasks $\{\mathcal{T}_1, \mathcal{T}_2, ..., \mathcal{T}_m\}$ arriving sequentially in the FL setup. Each task $\mathcal{T}_t$ contains data $\mathbf{D}_t$, at multiple clients, and data of task $t$ at client $c$ is defined as:

\begin{equation}
    \mathbf{D}_t^c = \{x_t^c(i), y_t^c(i)\}_{i=1}^{N_t^c}
\end{equation}

The tasks arrive at the server incrementally. Each time task $t$ arrives, the clients concerned with the task generate category-specific embeddings for each category $k\in\mathcal{Y}_t^c$.
\begin{equation}
    e_{t,k}^c(i) = \text{CLIP}_{\text{TEXT}}(\text{GPT-ViT}(x_{t,k}^c(i))) 
\end{equation}
where $e_{t,k}^c $ is the embedding generated by client $c$ of category $k$ in task $t$. GPT-ViT is a VLM used to generate textual descriptions for each sample $x_{t,k}^c(i)$, which is later encoded by CLIP~\cite{radford2021learning}. CLIP encodes each textual description into a 512-dimensional vector embedding. The image-specific embeddings are then averaged to calculate category-specific embeddings as:
\begin{equation}
    \mu_{t,k}^c = \frac{1}{N_{t,k}^c} \sum_i e_{t,k}(i)
\end{equation}
where $N_{t,k}^c$ is the number of images concerning category $k$, and task $t$ at client $c$.

CLIP embeddings organize semantic content linearly, so class means align well with image features, and can provide enough information about the actual data~\cite{radford2021learning}. The average CLIP embeddings category for each category is then sent to the server, significantly reducing the communication size compared to exchanging parameters. Furthermore, clients will communicate these embeddings only once. 

The server is equipped with a pretrained diffusion model $p_\phi(x|\mu)$ conditioned on $\mu$. The server uses the client-specific embeddings to generate new data for each category in the task $t$. The synthetic data generated by the diffusion model for class $k$ and task $t$ is denoted as:

\begin{equation}
    \hat{\mathbf{D}}_{t,k} = \{x_{t,k}(i)\}_{i=1}^z
\end{equation}
where $z$ is the number of samples generated for each category $k$ in task $t$.

Based on CLIP encodings aligning well with image features, and diffusion models' generalizable zero-shot ability~\cite{rombach2022high}, the synthesized data distribution resembles the actual data distribution. 

In a naive setup, as tasks and their data arrive incrementally at the server, a straightforward baseline is to fine-tune the model on each new task dataset as it arrives, without revisiting the earlier tasks. Formally, if $\hat{\mathbf{D}}_{t}$ denotes the dataset for task $t$, then at round $t$ the parameters are updated as:
\begin{equation}
\label{eq:naive-inc}
\theta_t
\;=\;
\arg\min_{\theta}\;
\frac{1}{\lvert \hat{\mathbf{D}}_{t}\rvert}
\sum_{(x,y)\in \hat{\mathbf{D}}_{t}}
\ell\bigl(f_{\theta}(x),y\bigr)
\quad
\end{equation}
While computationally cheap, this \emph{naive incremental} strategy leads to severe catastrophic forgetting: updates driven solely by $\hat{\mathbf{D}}_{t}$ overwrite knowledge useful for earlier tasks, resulting in significant performance degradation on past tasks.

When all tasks are centrally available (no incremental constraint), the OSCAR--style centralized training instead optimizes jointly across all task datasets:
\begin{equation}
\hat{\theta}
= \arg\min_{\theta}
\sum_{t=1}^{m}
\frac{1}{|\hat{\mathbf{D}}_t|}
\sum_{(x,y)\in \hat{\mathbf{D}}_t}
\ell\bigl(f_{\theta}(x),y\bigr)
\end{equation}
As the data arrives incrementally, OSCAR needs to retrain the model whenever new task data becomes available or its performance deteriorates, as we will see later in the experiments. Hence, the proposed approach, OSI-FL, incorporates a trade-off between retraining and performance. To alleviate forgetting without full retraining, OSI-FL retains $p$ exemplar samples \textit{per class} from past tasks for replay. Rather than selecting exemplars at random, OSI-FL employs an \textit{importance--based} criterion driven by gradient magnitude. For each $(x,y)\in \hat{\mathbf{D}}_t$, the importance score at parameters $\theta_{t-1}$ is given by:
\begin{equation}
    s_i(x,y;\theta_{t-1})\;=\;\bigl\|\nabla_{\theta}\,\ell\bigl(f_{\theta}(x),y\bigr)\bigr\|_2
\end{equation}

For each class $k$ in task $t$, we select the exemplar set $\mathcal{E}_{t,k}$ by
\begin{equation}
\label{eq:grad-sel-class}
\mathcal{E}_{t,k}
\;=\;
\underset{\substack{\mathcal{S}\subseteq \hat{\mathbf{D}}_{t,k}\\ \lvert \mathcal{S}\rvert=p}}{\arg\max}\;
\sum_{(x,y)\in \mathcal{S}} s_i(x,y;\theta_{t-1})
\end{equation}
where $\hat{\mathbf{D}}_{t,c}$ denotes the samples from class $k$ of task $t$. The final exemplar memory for task $t$ is then
\begin{equation}
\mathcal{E}_t \;=\; \bigcup_{k} \mathcal{E}_{t,k}
\end{equation}

This class-balanced selection ensures that all courses from past tasks are fairly represented in the replay buffer while still prioritizing the most informative samples within each class. Then, in each round when task $t$ arrives, OSI-FL incorporates reusing p-samples for each past task category to enable the previous tasks' knowledge retention, without a complete retraining on the data. In theory, OSI-FL solves:
\begin{equation}
\begin{split}
\theta_{t}
& = \arg\min_{\theta}
\Biggl[
\underbrace{\frac{1}{\lvert \hat{\mathbf{D}}_{t}\rvert}
\sum_{(x,y)\in \hat{\mathbf{D}}_{t}}\ell\bigl(f_{\theta}(x),y\bigr)}_{\text{new task}}\\
& +
\sum_{i=1}^{t-1}
\Bigl(
\underbrace{\frac{1}{\lvert \mathcal{E}_{i}\rvert}
\sum_{(x,y)\in \mathcal{E}_{i}}\ell\bigl(f_{\theta}(x),y\bigr)}_{\substack{\text{retained samples}}}
\Bigr)
\Biggr]
\end{split}
\end{equation}
where $\hat{\mathbf{D}}_{t}$ is the synthetic data for the current domain $t$, $\mathcal{E}_i$ are retained samples for previous domain $i$. The retained sample for each round is incorporated from sample retention in centralized incremental learning ~\cite{castro2018end}, where in each training round on the current task, we highlight the p-important samples to be retained for retraining in the next rounds, effectively eliminating the need for keeping and retraining the model with all samples and enabling knowledge retention from previous tasks. The sample selection strategy has been kept simple, as in each training round, the samples are sorted based on their training loss, and the top-p samples with the highest gradient magnitude values are added to the memory for training in the next rounds. A simplified overview of the proposed approach, OSI-FL, is depicted in Figure \ref{fig:osifl}.

\begin{table}[htbp]
  \centering    
    \caption{Hyperparameters}
    \begin{tabularx}{0.75\linewidth}{XX}
      \toprule
      Hyperparameter & Value \\
      \midrule
      Backbone & ResNet-18 \\
      Optimizer & Adam \\
      Learning rate   & 0.001   \\
      Batch size      & 256     \\
      Global epochs    & 20      \\
      Weight decay    & 1e-4   \\
      $p$  & 5\\
      \bottomrule
    \end{tabularx}
    \label{tab:hyperparameters}
    
  \end{table}%

  \begin{table}
  \centering
    \caption{Baselines}
    \begin{tabularx}{0.75\linewidth}{XMMM}
      \toprule
      Model           & FL            & OS     & IL       \\
      \midrule
      FedAvg & \greencheck   & \redcross     & \redcross\\
      FedProx         & \greencheck   & \redcross   & \redcross  \\
      FedDyn         & \greencheck   & \redcross   & \redcross  \\
      FedET         & \greencheck   & \redcross   & \greencheck  \\
      FedIL+         & \greencheck   & \redcross   & \greencheck  \\
      FedEWC      & \greencheck   & \redcross   & \greencheck  \\
      OSCAR           & \greencheck   & \greencheck  & \redcross \\
      OSCAR-R     & \greencheck   & \greencheck   & \greencheck \\
      OSCAR IL     & \greencheck   & \greencheck   & \greencheck \\
      \hline
      \rowcolor{lightgray}
      OSI-FL    & \greencheck   & \greencheck   & \greencheck \\
        
      \bottomrule
    \end{tabularx}
    \label{tab:modelperformance}
\end{table}

\section{Experimental Setup}\label{expsetup}
\subsubsection{Datasets} The experiments are carried out on three datasets - NICO++ Unique (NICO\_U), NICO++ Common (NICO\_C)~\cite{he2021towards}, and OpenImage~\cite{kuznetsova2020open}. Both NICO\_U and NICO\_C datasets have 60 categories over six domains. In NICO\_C, all the categories share the same domains, while in NICO\_U, the domains are different for every category. For the OpenImage dataset, we follow a similar setup as ~\cite{feddeo}, where 20 super categories are selected, and each super category contains six sub-categories, treated as domains. 

The experiments use two different setups: class incremental and domain incremental. In the class incremental setup, the tasks were assigned random categories from the data: ten categories each in the NICO\_U and NICO\_C experiments and three or four categories in the OpenImage experiments. All the experiments were conducted with six tasks for both domain incremental and class incremental setups, and every client was assigned a single task.

\subsubsection{Hyperparameters} As shown in Table \ref{tab:hyperparameters}, the learning rate for all the experiments was set to $0.001$, with the Adam optimizer, and ResNet-18 backbone as the feature extractor, which was frozen. The batch size is 32, and the number of global communication rounds for non-OSFL approaches is 20, with local training epochs set to 1. The value of the regularization parameter $\lambda$ for FedEWC and OSCAR-R is set to $0.1$. Furthermore, to keep the computational costs comparable, all the experiments are conducted with 50 images per client per category in all the datasets. In OSI-FL, the number of retained samples for every previous task, $p$, is set to 5 for every category. Also, to ensure result consistency and reproducibility, seed values of 42, 18, and 50 are used, and the results depicted in tables are an average of all three runs.

\subsubsection{Baselines} 
Due to the lack of existing literature in one-shot incremental FL, we compare OSI-FL to traditional FL algorithms - FedAvg, and FedProx, the incremental counterpart of FedAvg with regularization (FedEWC), and state-of-the-art OSFL approaches. In the FL approaches, the clients are incrementally trained and aggregated. This means that once task 2 arrives, the training will only happen on the client related to task 2, and aggregated to the global model of task 1. As OSCAR~\cite{oscar} uses a similar pipeline to OSI-FL, we use OSCAR, which is retrained on the complete data every time a new task is available, as the ceiling for the OSI-FL performance. We then devise different incremental variants of OSCAR. OSCAR-IL only trains on the newly arrived task, and OSCAR-R adds model-based regularization to OSCAR-IL. On the federated incremental learning front, OSI-FL is compared against FL with elastic transfer (FedET)~\cite{guerdan2023federated}, and FedIL+. Table \ref{tab:modelperformance} provides a detailed overview of the baselines. The official implementation of OSCAR is used to reproduce the results and develop its incremental variants\footnote{https://github.com/obaidullahzaland/oscar}.

\section{Results and Analysis}
\subsubsection{Main Results}
OSI-FL is compared against baselines in both class-incremental and domain-incremental setups, as described in section \ref{expsetup}. As shown in Table \ref{tab:labelincremental}, the performance of FL algorithms and the incremental variants of OSCAR is unpredictable in class incremental experiments. While in NICO\_C and NICO\_U, FL approaches perform better, in the OpenImage dataset, the OSCAR incremental variants are better. The difference can be related to NICO domains containing the same classes, while the OpenImage domains contain sub-classes that vary across domains (as they are hand-crafted). OSI-FL performs better than the incremental variants of both FL (FedEWC and FedIL+), and OSCAR (OSCAR-R and OSCAR-IL). The incremental versions of OSCAR experience a significant drop in performance across both variations of the NICO++ dataset, while the incremental FL algorithms struggle on the OpenImage datasets. OSI-FL performs more linearly, while retraining with only $p=5$ samples per class from the previous tasks in each round.

\begin{table}[!htb]
    \centering
    \caption{Accuracy (in \%) on the test set for the baselines and our approach on class incremental settings after each task. Ceiling performance in italic, best performance excluding ceiling in bold.}
    \begin{tabularx}{1\linewidth}{m{1.3cm}*{6}{>{\centering\arraybackslash}X}}
    \hline
         \textbf{Model} &  \multicolumn{6}{c}{\textbf{Client Test Set Accuracy after each task}}$\uparrow$ \\
         \hline
         
         \hline
         & $\mathcal{T}_1$ &   $\mathcal{T}_2$ & $\mathcal{T}_3$ & $\mathcal{T}_4$  & $\mathcal{T}_5$ &  $\mathcal{T}_6$ \\
        \hline
            \multicolumn{7}{c}{\textbf{OpenImage}}\\
         \hline

         \hline
         FedAvg & 93.69 & 41.54 & 33.97 & 28.94 & 24.75 & 25.22\\
         FedProx & 93.69 & 41.54 & 33.98 & 29.64 & 24.73 & 25.22  \\
         \hline
         FedEWC & 93.69 & 41.64 & 34.22 & 29.02 & 25.02 & 25.19 \\
         
         FedET & 95.22 & 44.52 & 39.94 & 34.14 & 33.61 & 32.27  \\
         FedIL+ & 63.24 & 49.87 & 49.26 & 47.26 & 47.92 & 44.56  \\
         \hline
         OSCAR-IL & 53.50 & 47.50 & 48.33 & 45.85 & 46.76 & 45.76 \\
         OSCAR-R & 53.50 & 45.41 & 47.56 & 44.64 & 45.42 & 45.24  \\
         \rowcolor{lightgray}
         OSI-FL  & 53.50 & \textbf{52.36} & \textbf{53.85} & \textbf{57.18} &\textbf{60.85} & \textbf{56.67} \\
         \hline
            \multicolumn{7}{c}{\textbf{NICO\_U}}\\
         \hline

         \hline
         FedAvg & 96.99 & 64.17 & 50.05 & 42.82 & 49.98 & 39.86\\
         FedProx & 96.99 & 64.22 & 50.05 & 42.82 & 49.98 & 39.83  \\
         \hline
         FedEWC & 96.09 & 64.22  & 50.43 & 42.69 & 50.80 & 40.09 \\
         
         FedET & 98.03 & 62.85 & 51.71 & 43.24 & 47.98 & 43.62  \\
         FedIL+ & 92.66 & 66.58 & 63.14 & 54.76 & 51.32 & 49.98  \\
         \hline
         OSCAR-IL & 89.96 & 50.79 & 32.54 & 33.34 & 28.90 & 25.96  \\
         OSCAR-R & 89.96 & 50.35 & 31.64 & 33.58 & 28.94 & 26.03  \\
         \rowcolor{lightgray}
         OSI-FL  & 89.96 & \textbf{72.67} & \textbf{70.38} & \textbf{65.52} & \textbf{58.68} & \textbf{58.88} \\

         \hline

         \multicolumn{7}{c}{\textbf{NICO\_C}}\\
         \hline

            \hline
         FedAvg & 89.03 & 51.68 & 40.75 & 32.19 & 38.01 & 30.56 \\
         FedProx & 88.96 & 51.61 & 40.70 & 32.19 & 37.99 & 30.58  \\
         \hline
         FedEWC & 88.96 & 51.85 & 40.68 & 32.33 & 37.82 & 30.52 \\
         
         FedET & 90.38 & 52.85 & 41.71 & 33.24 & 37.98 & 31.62 \\
         FedIL+ & 90.16 & 56.33 & 49.77 & 46.82 & 45.26 & 41.88  \\
         \hline
         OSCAR-IL & 79.67 & 35.45 & 23.97 & 28.02 & 23.84 & 22.45 \\
         OSCAR-R & 79.67 & 35.67 & 23.57 & 28.51 & 23.90 & 22.41  \\
         \rowcolor{lightgray}
         OSI-FL  & 79.67 & \textbf{60.39} & \textbf{59.43} & \textbf{54.96} & \textbf{48.25} & \textbf{49.76} \\

         \hline

        \hline
    \end{tabularx}
    \label{tab:labelincremental}
\end{table}

\begin{table}[!htb]
    \centering
    \caption{Accuracy (in \%) on the test set for the baselines and our approach on domain incremental settings after each task. Ceiling performance in italic, best performance excluding ceiling in bold}
    \begin{tabularx}{1\linewidth}{m{1.3cm}*{6}{>{\centering\arraybackslash}X}}
    \hline
         \textbf{Model} &  \multicolumn{6}{c}{\textbf{Client Test Set Accuracy after each task}}$\uparrow$ \\
         \hline
         
         \hline
         
         & $\mathcal{T}_1$ &   $\mathcal{T}_2$ & $\mathcal{T}_3$ & $\mathcal{T}_4$  & $\mathcal{T}_5$ &  $\mathcal{T}_6$ \\
         \hline

        \hline
            \multicolumn{7}{c}{\textbf{OpenImage}}\\
         \hline

         \hline

         FedAvg & 64.38 & 52.64 & 53.08 & 58.44 &58.68  &57.14 \\
         FedProx & 64.38 & 52.55 & 53.62 & 53.94 & 57.25 & 57.20  \\
         FedEWC & \textbf{64.38} & \textbf{52.72} & 53.42 & \textbf{59.66} & 59.22 &58.04  \\
         \hline
         OSCAR-IL & 52.74 & 47.40 & 48.72 & 45.66 & 46.67 & 45.78\\
         OSCAR-R & 52.74 & 45.41 & 47.56 & 44.64 & 45.42 & 45.24  \\
         \rowcolor{lightgray}
         OSI-FL  & 52.74 & 52.36 & \textbf{53.85} & 52.74& \textbf{60.85} & \textbf{59.67}  \\

         \hline
         
         \hline
            \multicolumn{7}{c}{\textbf{NICO\_U}}\\
         \hline

         \hline

         FedAvg & 69.22 & 68.25 & 72.87 & 72.16 & 73.27 & 73.15\\
         FedProx & 69.22 & 68.04 & 72.96 & 73.22 & 71.47 & 73.09  \\
         FedEWC & \textbf{69.22} & 68.77 & 73.24 & 73.96 & 73.84 &73.33  \\
         \hline
         OSCAR-IL & 59.98 & 68.35 & 70.28 & 69.24 & 70.22 & 70.26  \\
         OSCAR-R & 59.98 & 66.22 & 67.51 & 67.03 & 67.70 & 68.11  \\
         \rowcolor{lightgray}
         OSI-FL  & 59.98 & \textbf{69.39} &\textbf{73.50} & \textbf{73.46} & \textbf{75.75} & \textbf{73.85} \\

         \hline

         \hline

         \multicolumn{7}{c}{\textbf{NICO\_C}}\\
         \hline

         \hline


         FedAvg & 67.83 & 61.98 & 56.66 & 58.48 & 57.46 & 54.17 \\
         FedProx & 67.83 & 61.83 & 55.63 & 58.48 & 54.92 & 53.66  \\
         FedEWC & \textbf{67.83} & 61.98 & 57.92 & 59.74 & 59.15 & 56.44 \\
         \hline
         OSCAR-IL & 61.78 & 58.26 & 53.72 & 54.03 & 56.16 & 53.87  \\
         OSCAR-R & 61.78 & 59.20 & 53.56 & 54.95 & 56.12 & 54.24 \\
         \rowcolor{lightgray}
         OSI-FL  & 61.78 & \textbf{62.50} & \textbf{58.25} & \textbf{60.72} & \textbf{62.26} & \textbf{60.19}  \\
         \hline

         \hline

    \end{tabularx}
    \label{tab:domainincremental}
\end{table}


In the domain incremental experiments, as shown in Table \ref{tab:domainincremental}, a similar story unfolds. However, the incremental versions of OSCAR (OSCAR-IL and OSCAR-R) and OSI-FL. This is because now each task contains all the classes, making the tasks easier than a class-based incremental setup. OSI-FL still improves on both the incremental versions of OSCAR and the traditional FL algorithms (FedAvg and FedProx). The domain incremental results show that OSI-FL can approach ceiling performance, and, with a higher value of $p$ and more robust sample selection approaches, may potentially reach it with a fraction of the computational cost.

\begin{figure}
    \centering
    \includegraphics[width=\linewidth]{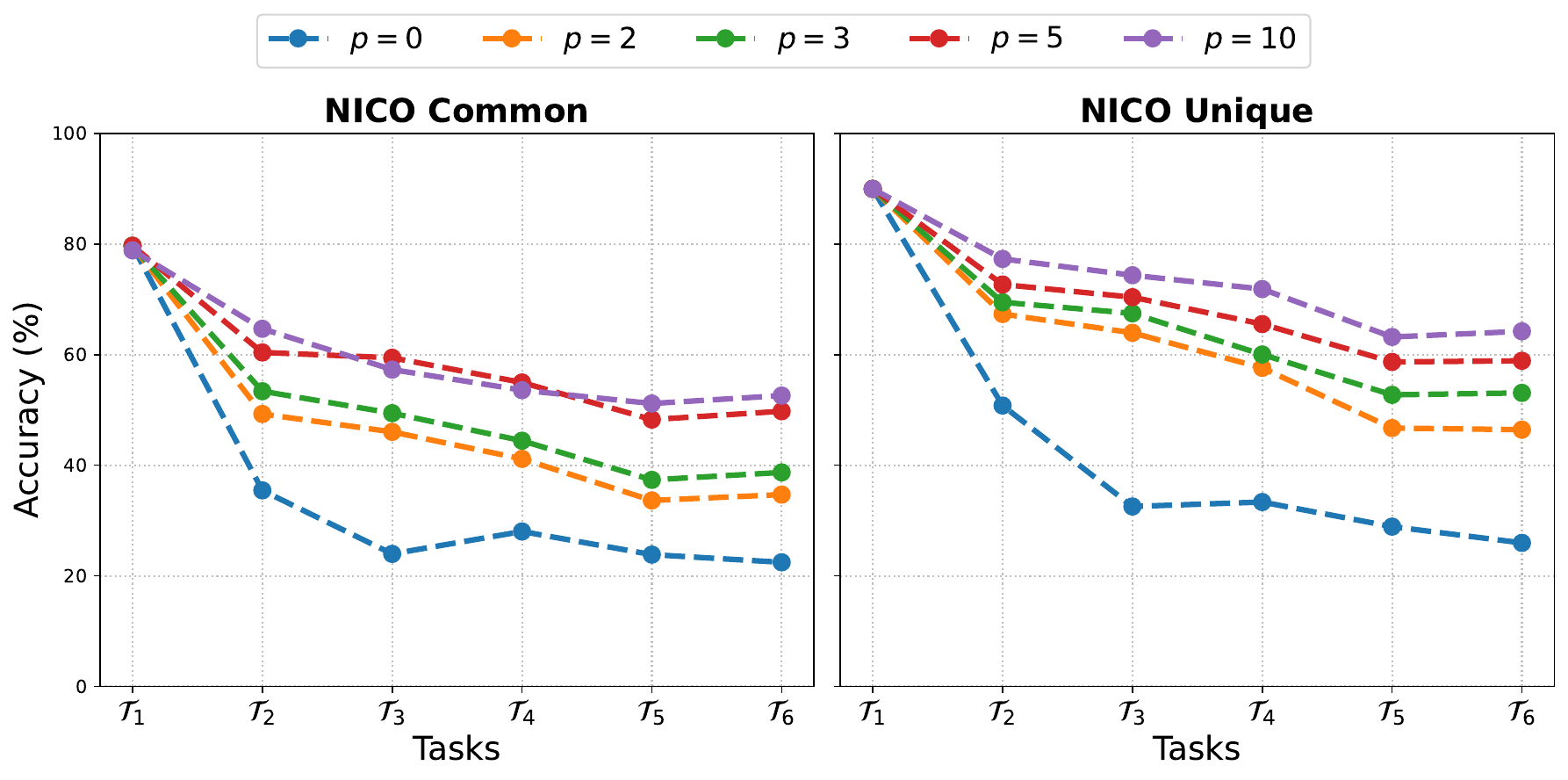}
    \caption{Impact of $p$, the number of retained samples, on the overall performance in OSI-FL.}
    \label{fig:replay}
\end{figure}

\subsubsection{Ablation Study}
OSI-FL is dependent on the value of $p$ - the number of samples to be retained from each class and task pair. As can be seen in \textit{Fig.} \ref{fig:replay}, even retraining two samples per class and task pair improves performance on OSCAR-IL, where no samples from previous domains are retained. The most significant jump in overall performance is observed as the value of $p$ changes from 0 to 2, with around 20\% overall performance improvement in NICO\_U and around 15\% overall accuracy improvement in NICO\_C. While the increase in the number of retained samples is proportional to the improvement in performance, it also adds overall computational and memory costs.

\subsubsection{Number of Clients}

\begin{figure}
    \centering
    \includegraphics[width=\linewidth]{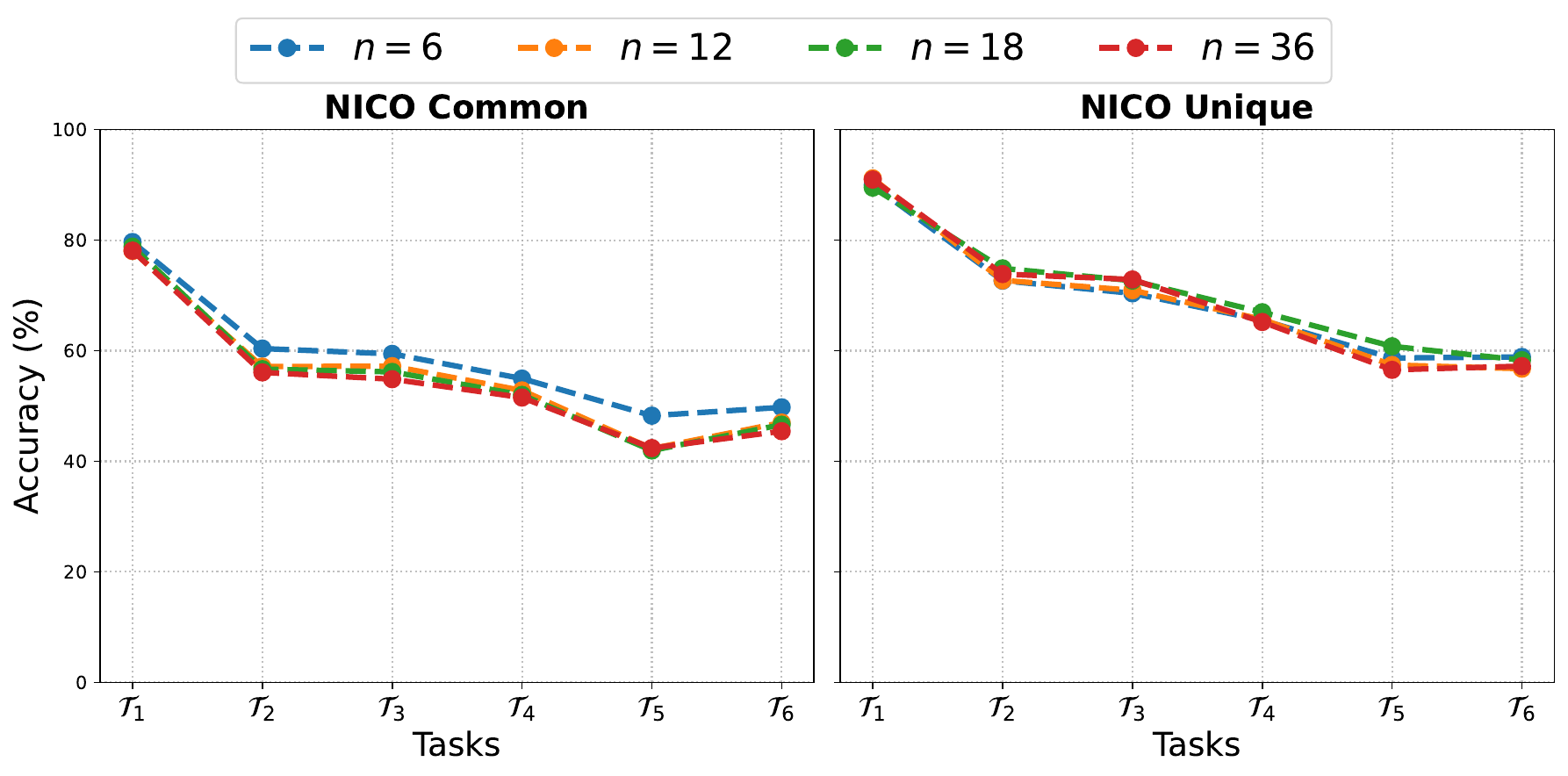}
    \caption{Class incremental accuracy with varying number of clients}
    \label{fig:clients}
\end{figure}

OSI-FL provides a scalable solution, as shown in \textit{Fig.} \ref{fig:clients}, increasing the number of clients from 6 to 36 minimally affects the performance of the global model. This indicates that the framework is robust to the heterogeneity introduced by larger client participation, ensuring that the collaborative training process does not compromise accuracy. Moreover, the stability of performance across varying client scales highlights the efficiency of OSI-FL in managing communication overhead while maintaining consistent generalization ability. Such scalability is crucial for real-world FL scenarios, where the number of participating devices can fluctuate significantly.

\begin{figure}
    \centering
    \includegraphics[width=\linewidth]{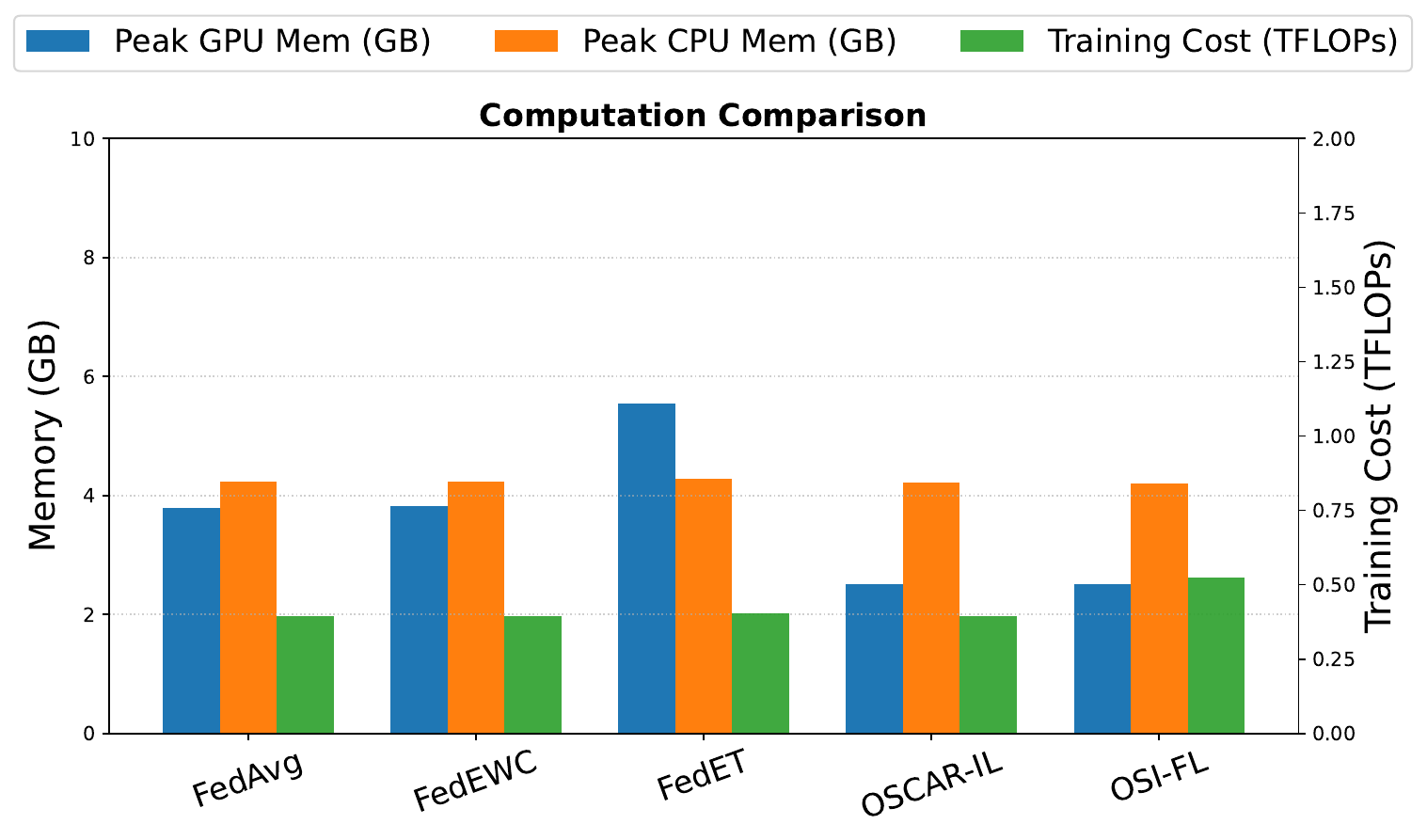}
    \caption{Peak GPU and CPU memory usage (in GB), and training cost (in TFLOPS) for OSI-FL and baselines (NICO Common dataset and ResNet-18 backbone). }
    \label{fig:computation}
\end{figure}

\begin{figure}
    \centering
    \includegraphics[width=0.9\linewidth]{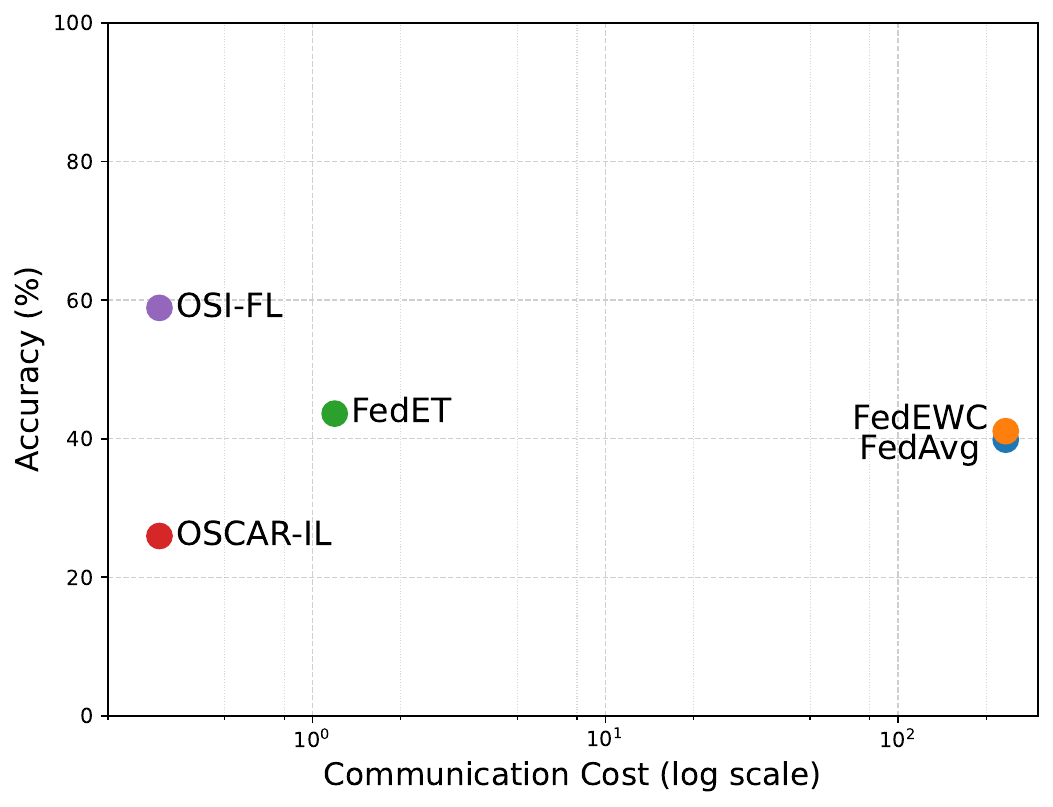}
    \caption{Client upload cost in OSI-FL and baselines compared to final performance on class-incremental OpenImage dataset and ResNet-18 backbone (in million parameters, log scaled).}
    \label{fig:communication}
\end{figure}

\subsubsection{Computation and Communication Cost}
\textit{Fig.} \ref{fig:computation} shows the computation cost of OSI-FL against baseline approaches. While it is evident that OSI-FL increases the computational cost marginally (as it trains on retained samples from the previous tasks), the increment in computational cost is not high. On the other hand, OSI-FL, alongside its one-shot FL counterpart OSCAR-IL, has the lowest peak GPU footprint, utilizing around 2 GB of GPU memory at its peak. Traditional FL algorithms, however, use at least twice the GPU memory, with FedET having the highest footprint at around 6 GBs. All the baselines have equivalent peak CPU memory footprint. 

On the communication front, one-shot FL approaches - OSCAR-IL and OSI-FL have similar client upload cost, as both upload only category-specific data encodings, which is at least 5 times lower than FedET. Traditional FL algorithms - such as FedAvg - on the other hand, have the highest client upload cost at 233 million parameters per client (considering ResNet-18 and 20 global epochs). \textit{Fig.} \ref{fig:communication} shows the log-scaled uploaded parameters by each client against their performance on the OpenImage dataset. 

\section{Conclusion}
This work introduces OSI-FL, a first one-shot incremental federated learning framework that utilizes pretrained foundation and diffusion models to reduce the overall communication rounds in IFL to just one. OSI-FL further incorporates selective sample retention (SSR) to select top-p samples for each task-class pair for retraining in subsequent rounds, thereby tackling catastrophic forgetting in IFL. The experiments across three datasets, in both class- and domain-incremental setups, indicated that OSI-FL achieves superior performance compared to the incremental FL variants alongside incremental variants of the existing OSFL framework.

While the existing OSI-FL framework acts as a pioneering work in one-shot incremental federated learning, it holds the potential to improve in certain aspects, most notably the sample selection for retention. The existing framework uses a naive gradient-magnitude approach, but more robust sample retention strategies can enable OSI-FL to select diverse samples that retain higher knowledge from previous tasks. Additionally, integrating local models' knowledge can potentially improve the global model's performance.

\bibliography{main}

@inproceedings{mcmahan2017communication,
  author       = {Brendan McMahan and
                  Eider Moore and
                  Daniel Ramage and
                  Seth Hampson and
                  Blaise Ag{\"{u}}era y Arcas},
  title        = {Communication-Efficient Learning of Deep Networks from Decentralized
                  Data},
  booktitle    = {Proceedings of the 20th International Conference on Artificial Intelligence
                  and Statistics, {AISTATS} 2017, 20-22 April 2017, Fort Lauderdale,
                  FL, {USA}},
  series       = {Proceedings of Machine Learning Research},
  volume       = {54},
  pages        = {1273--1282},
  publisher    = {{PMLR}},
  year         = {2017},
  bibsource    = {dblp computer science bibliography, https://dblp.org}
}

@inproceedings{oscar,
  author={Zaland, Obaidullah and Jin, Shutong and Pokorny, Florian T. and Bhuyan, Monowar},
  booktitle={2025 IEEE International Conference on Multimedia and Expo (ICME)}, 
  title={One-Shot Federated Learning with Classifier-Free Diffusion Models}, 
  year={2025},
  volume={},
  number={},
  pages={1-6},
  doi={10.1109/ICME59968.2025.11209111}
}

@inproceedings{radford2021learning,
  title={Learning transferable visual models from natural language supervision},
  author={Radford, Alec and Kim, Jong Wook and Hallacy, Chris and Ramesh, Aditya and Goh, Gabriel and Agarwal, Sandhini and Sastry, Girish and Askell, Amanda and Mishkin, Pamela and Clark, Jack and others},
  booktitle={International conference on machine learning},
  pages={8748--8763},
  year={2021},
  organization={PmLR}
}

@inproceedings{rombach2022high,
  title={High-resolution image synthesis with latent diffusion models},
  author={Rombach, Robin and Blattmann, Andreas and Lorenz, Dominik and Esser, Patrick and Ommer, Bj{\"o}rn},
  booktitle={Proceedings of the IEEE/CVF conference on computer vision and pattern recognition},
  pages={10684--10695},
  year={2022}
}

@article{chen2024federated,
  title={When federated learning meets privacy-preserving computation},
  author={Chen, Jingxue and Yan, Hang and Liu, Zhiyuan and Zhang, Min and Xiong, Hu and Yu, Shui},
  journal={ACM Computing Surveys},
  volume={56},
  number={12},
  pages={1--36},
  year={2024},
  publisher={ACM New York, NY}
}

@article{li2020federated,
  title={Federated optimization in heterogeneous networks},
  author={Li, Tian and Sahu, Anit Kumar and Zaheer, Manzil and Sanjabi, Maziar and Talwalkar, Ameet and Smith, Virginia},
  journal={Proceedings of Machine learning and systems},
  volume={2},
  pages={429--450},
  year={2020}
}

@inproceedings{karimireddy2020scaffold,
  title={Scaffold: Stochastic controlled averaging for federated learning},
  author={Karimireddy, Sai Praneeth and Kale, Satyen and Mohri, Mehryar and Reddi, Sashank and Stich, Sebastian and Suresh, Ananda Theertha},
  booktitle={International conference on machine learning},
  pages={5132--5143},
  year={2020},
  organization={PMLR}
}

@article{osfl,
  title={One-shot federated learning},
  author={Guha, Neel and Talwalkar, Ameet and Smith, Virginia},
  journal={arXiv preprint arXiv:1902.11175},
  year={2019}
}

@article{dosfl,
  title={Distilled one-shot federated learning},
  author={Zhou, Yanlin and Pu, George and Ma, Xiyao and Li, Xiaolin and Wu, Dapeng},
  journal={arXiv preprint arXiv:2009.07999},
  year={2020}
}

@inproceedings{feddeo,
  title={FedDEO: Description-Enhanced One-Shot Federated Learning with Diffusion Models},
  author={Yang, Mingzhao and Su, Shangchao and Li, Bin and Xue, Xiangyang},
  booktitle={Proceedings of the 32nd ACM International Conference on Multimedia},
  pages={6666--6675},
  year={2024}
}

@article{dense,
  title={Dense: Data-free one-shot federated learning},
  author={Zhang, Jie and Chen, Chen and Li, Bo and Lyu, Lingjuan and Wu, Shuang and Ding, Shouhong and Shen, Chunhua and Wu, Chao},
  journal={Advances in Neural Information Processing Systems},
  volume={35},
  pages={21414--21428},
  year={2022}
}

@article{ifl,
  title={No one left behind: Real-world federated class-incremental learning},
  author={Dong, Jiahua and Li, Hongliu and Cong, Yang and Sun, Gan and Zhang, Yulun and Van Gool, Luc},
  journal={IEEE Transactions on Pattern Analysis and Machine Intelligence},
  volume={46},
  number={4},
  pages={2054--2070},
  year={2023},
  publisher={IEEE}
}

@article{iflmodel,
  author={Wu, Feng and Ziying Tan, Alysa and Feng, Siwei and Yu, Han and Deng, Tao and Zhao, Libang and Chen, Yuanlu},
  journal={IEEE Internet of Things Journal}, 
  title={Federated Class-Incremental Learning via Weighted Aggregation and Distillation}, 
  year={2025},
  volume={12},
  number={12},
  pages={22489-22503},
  doi={10.1109/JIOT.2025.3553901}}

@article{iflmodel2,
  title={Cyclical Weight Consolidation: Towards Solving Catastrophic Forgetting in Serial Federated Learning},
  author={Song, Haoyue and Wang, Jiacheng and Wang, Liansheng},
  journal={arXiv preprint arXiv:2405.10647},
  year={2024}
}

@inproceedings{ifldata,
  author={Li, Yichen and Li, Qunwei and Wang, Haozhao and Li, Ruixuan and Zhong, Wenliang and Zhang, Guannan},
  booktitle={2024 IEEE/CVF Conference on Computer Vision and Pattern Recognition (CVPR)}, 
  title={Towards Efficient Replay in Federated Incremental Learning}, 
  year={2024},
  volume={},
  number={},
  pages={12820-12829},
  doi={10.1109/CVPR52733.2024.01218}}

@article{ifldata2,
author={Li, Yichen and Wang, Haozhao and Qi, Yining and Liu, Wei and Li, Ruixuan},
journal={ IEEE Transactions on Pattern Analysis \& Machine Intelligence },
title={{ Re-Fed+: A Better Replay Strategy for Federated Incremental Learning }},
year={2025},
volume={47},
number={07},
ISSN={1939-3539},
pages={5489-5500},
doi={10.1109/TPAMI.2025.3551732},
publisher={IEEE Computer Society},
address={Los Alamitos, CA, USA},
month=jul}

@article{ifldomain,
author = {Li, Yichen and Xu, Wenchao and Qi, Yining and Wang, Haozhao and Li, Ruixuan and Guo, Song},
title = {SR-FDIL: Synergistic Replay for Federated Domain-Incremental Learning},
year = {2024},
issue_date = {Nov. 2024},
publisher = {IEEE Press},
volume = {35},
number = {11},
issn = {1045-9219},
doi = {10.1109/TPDS.2024.3436874},
journal = {IEEE Trans. Parallel Distrib. Syst.},
month = nov,
pages = {1879–1890},
numpages = {12}
}

@inproceedings{iflclass,
  title={Federated class-incremental learning},
  author={Dong, Jiahua and Wang, Lixu and Fang, Zhen and Sun, Gan and Xu, Shichao and Wang, Xiao and Zhu, Qi},
  booktitle={Proceedings of the IEEE/CVF conference on computer vision and pattern recognition},
  pages={10164--10173},
  year={2022}
}

@inproceedings{iflparameter,
  title={Personalized Federated Class-Incremental Learning through Critical Parameter Transfer},
  author={Wu, Feng and Feng, Siwei and Chen, Yuanlu and Zhao, Libang},
  booktitle={ICASSP 2025-2025 IEEE International Conference on Acoustics, Speech and Signal Processing (ICASSP)},
  pages={1--5},
  year={2025},
  organization={IEEE}
}

@article{li2025challenges,
title = {From challenges and pitfalls to recommendations and opportunities: Implementing federated learning in healthcare},
journal = {Medical Image Analysis},
volume = {101},
pages = {103497},
year = {2025},
issn = {1361-8415},
doi = {https://doi.org/10.1016/j.media.2025.103497},
author = {Ming Li and Pengcheng Xu and Junjie Hu and Zeyu Tang and Guang Yang},
}

@article{fu2024secure,
  author={Fu, Yuchuan and Tang, Xinlong and Li, Changle and Yu, Fei Richard and Cheng, Nan},
  journal={IEEE Transactions on Intelligent Transportation Systems}, 
  title={A Secure Personalized Federated Learning Algorithm for Autonomous Driving}, 
  year={2024},
  volume={25},
  number={12},
  pages={20378-20389},
  doi={10.1109/TITS.2024.3450726}}

@article{casado2023ensemble,
  title={Ensemble and continual federated learning for classification tasks},
  author={Casado, Fernando E and Lema, Dylan and Iglesias, Roberto and Regueiro, Carlos V and Barro, Sen{\'e}n},
  journal={Machine Learning},
  volume={112},
  number={9},
  pages={3413--3453},
  year={2023},
  publisher={Springer}
}

@article{hamedi2025federated,
title = {Federated continual learning: Concepts, challenges, and solutions},
journal = {Neurocomputing},
volume = {651},
pages = {130844},
year = {2025},
issn = {0925-2312},
doi = {https://doi.org/10.1016/j.neucom.2025.130844},
author = {Parisa Hamedi and Roozbeh Razavi-Far and Ehsan Hallaji},
}

@inproceedings{mendieta2025navigating,
  title={Navigating heterogeneity and privacy in one-shot federated learning with diffusion models},
  author={Mendieta, Matias and Sun, Guangyu and Chen, Chen},
  booktitle={2025 IEEE/CVF Winter Conference on Applications of Computer Vision (WACV)},
  pages={2601--2610},
  year={2025},
  organization={IEEE}
}

@article{he2021towards,
  title={Towards non-iid image classification: A dataset and baselines},
  author={He, Yue and Shen, Zheyan and Cui, Peng},
  journal={Pattern Recognition},
  volume={110},
  pages={107383},
  year={2021},
  publisher={Elsevier}
}

@article{kuznetsova2020open,
  title={The open images dataset v4: Unified image classification, object detection, and visual relationship detection at scale},
  author={Kuznetsova, Alina and Rom, Hassan and Alldrin, Neil and Uijlings, Jasper and Krasin, Ivan and Pont-Tuset, Jordi and Kamali, Shahab and Popov, Stefan and Malloci, Matteo and Kolesnikov, Alexander and others},
  journal={International journal of computer vision},
  volume={128},
  number={7},
  pages={1956--1981},
  year={2020},
  publisher={Springer}
}

@inproceedings{castro2018end,
  title={End-to-end incremental learning},
  author={Castro, Francisco M and Mar{\'\i}n-Jim{\'e}nez, Manuel J and Guil, Nicol{\'a}s and Schmid, Cordelia and Alahari, Karteek},
  booktitle={Proceedings of the European conference on computer vision (ECCV)},
  pages={233--248},
  year={2018}
}

@article{iqbal2025hierarchical,
  author={Iqbal, Saeed and Zhong, Xiaopin and Khan, Muhammad Attique and Wu, Zongze and AlHammadi, Dina Abdulaziz and Liu, Weixiang and Choudhry, Imran Arshad},
  journal={IEEE Transactions on Consumer Electronics}, 
  title={Hierarchical Continual Learning for Domain-Knowledge Retention in Healthcare Federated Learning}, 
  year={2025},
  volume={71},
  number={2},
  pages={5025-5035},
  doi={10.1109/TCE.2025.3563909}}

@inproceedings{zalandiconip,
author="Zaland, Obaidullah
and Onur, Yakup
and Bhuyan, Monowar",
title="Mitigating Data Heterogeneity with Multi-tier Federated GAN",
booktitle="31st International Conference on Neural Information Processing (ICONIP)",
year="2024",
publisher="Springer Nature Singapore",
address="Singapore",
pages="225--239",
isbn="978-981-96-6972-1"
}

@article{yu2025efficient,
  title={Efficient Federated Class-Incremental Learning of Pre-Trained Models via Task-agnostic Low-rank Residual Adaptation},
  author={Yu, Feng and Hu, Jia and Min, Geyong},
  journal={arXiv preprint arXiv:2505.12318},
  year={2025}
}

@inproceedings{guerdan2023federated,
  title={Federated continual learning for socially aware robotics},
  author={Guerdan, Luke and Gunes, Hatice},
  booktitle={2023 32nd IEEE International Conference on Robot and Human Interactive Communication (RO-MAN)},
  pages={1522--1529},
  year={2023},
  organization={IEEE}
}

@article{amato2025towards,
  title={Towards One-shot Federated Learning: Advances, Challenges, and Future Directions},
  author={Amato, Flora and Qiu, Lingyu and Tanveer, Mohammad and Cuomo, Salvatore and Giampaolo, Fabio and Piccialli, Francesco},
  journal={arXiv preprint arXiv:2505.02426},
  year={2025}
}

\end{document}